# Object Recognition through Pose and Shape Estimation

*D. Anitta, School of Electronics Engineering, VIT University, Chennai, Tamilnadu, India. E-mail:anitta.d2015@vit.ac.in*

*A. Annis Fathima, School of Electronics Engineering, VIT University, Chennai, Tamilnadu, India.*

**Abstract**--- Computer vision helps machines or computer to see like humans. Computer Takes information from the images and then understands of useful information from images. Gesture recognition and movement recognition are the current area of research in computer vision. For both gesture and movement recognition finding pose of an object is of great importance. The purpose of this paper is to review many state of art which is already available for finding the pose of object based on shape, based on appearance, based on feature and comparison for its accuracy, complexity and performance

**Keywords**--- Object Recognition, Face Recognition, Face Orientation

## I. Introduction

A framework utilizing object detection strategy identifies the object through the comparison of object picture with the present reality objects. This becomes possible through object models by the prior information. Humans can identify the objects effortlessly. But it is very complex to execute this methodology on machines

The components required for completing the object detection task are: Model database, Feature detector, Hypothesizer and Hypothesis verifier. The flow is shown in Fig.1.

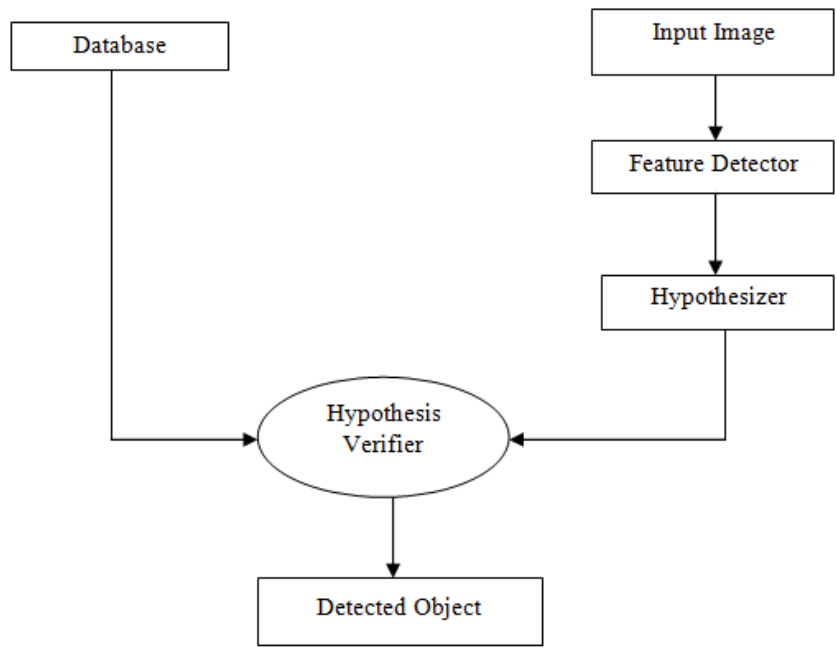

Figure 1: General flow of Object Detection

The database contains information about the models available to the system. The information in the database relies on the methodology utilized for the identification. Mostly, the object models are abstract feature vectors, as defined later in this article. An element is some trait of the item that is viewed as critical in estimating and identifying the object with respect to other objects. Size, colour, and shape are some examples of elements. The characteristics utilized by a framework rely upon the sorts of things to be identified and the association of the model-data-base. Utilizing the identified components as a part of the picture, the hypothesizer assigns probabilities to items present in the scene. This flow is utilized to lessen the searching space for the detector utilizing certain components. The verifier makes use of object models to confirm the speculations and filters the probability of items. The framework then chooses the object with the highest probability, on the basis of evidence, as the actual object.

## II. Applications of Object Recognition

- To find defective products in industry
- Face Recognition helps in security systems like Aadhar and Passport
- To find tumour growth in medical industry
- To find a person or an object from a photo
- To Count Objects in industries

The object recognition technique is processed by comparing the object with the stored image from the database. This comparison is carried out on the basis of shape, appearance and feature. Thus the estimation is categorized into 3 following categories

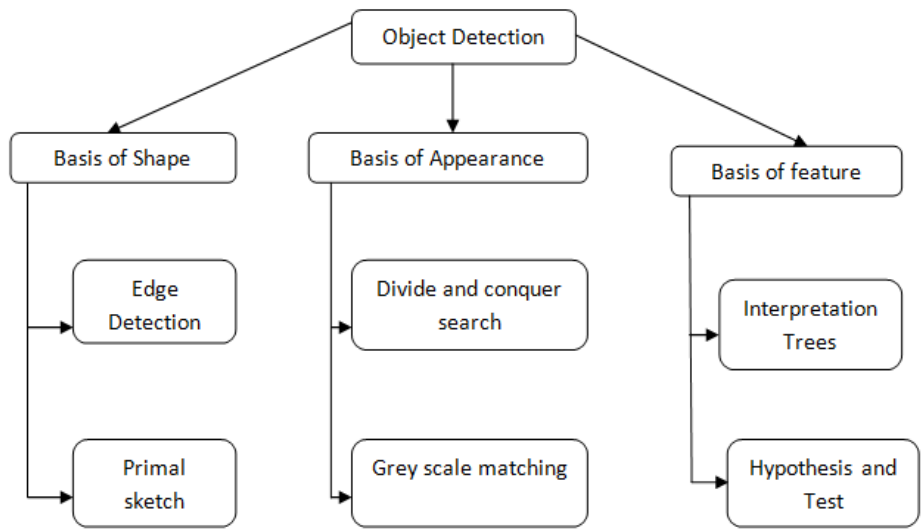

Figure 2: Types of Object Detection

## III. Estimation on Basis of Shape

Early endeavours on detection of objects were centred on utilizing shape based models of things to represent their appearance variety because of change in perspective. The principle of defining the shape of a 3D object which permits the prediction of elevated shape precisely in a 2D format, in this way encouraging detection mechanism utilizing edge or boundary data (which will not vary under changing circumstances). More concentration was given to shape aspects such as circle, square, line etc. which will not vary according to the changes in perspective. A phenomenal survey on object recognition on the basis of shape researched by Mundy can likewise be found in [3].Methods to find Real time shape deformation of the objects by using Interior Radial basis functions with reduced local distortions are discussed in [4].The use of Different trained human hand shapes and Possible object related to that hand shape is discussed in[5].Common base triangle area is another high accuracy technique, In which the CBTA of contour points are calculated and then the match cost matrix is calculated which help in finding the similarity of two shapes[6]

Table 1: Positives and Negatives of Shape based Methods

| Positives | Negatives |
|---|---|
| Invariance to illumination | problems of missing features |
| Invariance to view point | noisy geometry |
| Many mathematical formulas available to find shape | |

Even though this method has limitation, It was possible to recognize many of the 3D objects in the Early Years of Object Recognition.

## IV. Estimation on Basis of Appearance

Though many attempts has been made on shape-based object recognition works, latest endeavours have been focused on appearance-based strategies as progressive characteristic descriptors and on the basis of its appearance strategy is developed. Most strikingly, the Eigen face strategies have pulled in much consideration as it is one of the principal face detection frameworks that are computationally productive and generally precise. The fundamental thought of this methodology is to calculate Eigen vectors from an arrangement of vectors where everyone recognize one face picture as a raster scan vector of gray-scale pixel values. Each Eigen vector, named as an Eigen face, catches certain changes among every vectors, and a little set of Eigen vectors catches all the changes in the appearance of face pictures given in the set. If provided a test picture denoted by a vector of grey-scale pixel values, its character is identified by finding the closest neighbour of this vector after the projection on space traversed by a set of Eigen vectors. Every face picture can be depicted by a linear blend of Eigen faces with least mistake, and this linear mix constitutes a minimized reorientation. The Eigen face approach has been embraced in detecting objects crosswise over different views and displaying brightening changes [3].

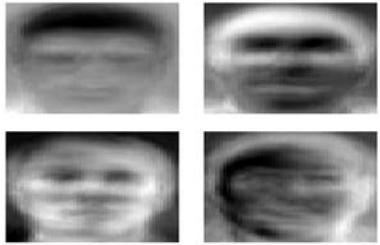

Figure 3: Eigen Faces

Another appearance based object detection is discussed in [7]. The paper uses various head pose descriptors and allows the head pose estimation on the basis of 3 features i.e. roll, yaw and pitch. The vertical variation of head position is described by pitch, the horizontal variation of head position is described by yaw and the inclination variation of head position is described by roll. Estimation of Pose and Direction of Multiple Fishes in a Aquarium on the basis of appearance model is said in Paper [8] in which the challenge of Occlusion is overcomed by knowing the number of fishes in the Aquarium. This method gives less than 4% of error. Paper[9] discuss about detection of movement of head pose by identifying the face(Object) and its movements Using an active appearance model, Here Roll, Pitch and Yaw angle are used to find the rotations of Head, Pitch, Row and Yaw Position are shown in figure 4.

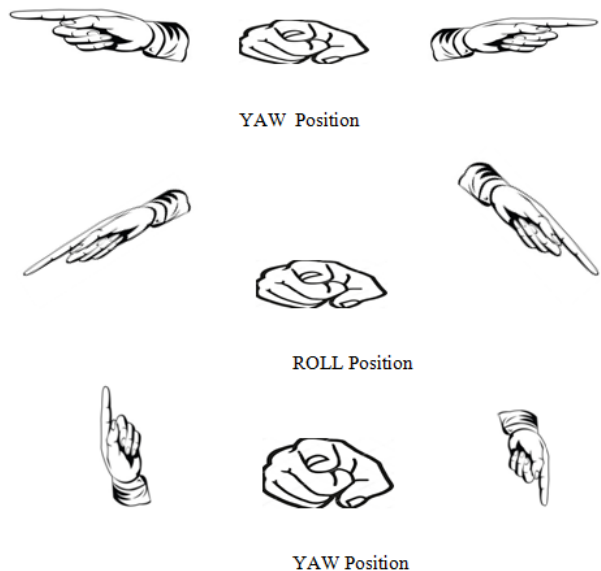

Figure 4: Pitch, Row and Yaw Position

Various Feature Detector methods and their repeatability, Run time etc are compared in Table 2.

Table 2: Comparison of Some Famous Feature Detectors on the Basis of Appearance

| Detector | Assigned Category | Invariance | Run Time | Repeatability | Nuumber of detections |
|---|---|---|---|---|---|
| Harris | Corner | None | Very Short | High | High |
| Hessian | Region | None | Very Short | High | High |
| DoG | Region | Scale | Short | High | Medium |
| MSER | Region | Projective | Short | High | Low |

## V. Estimation on Basis of Feature

The focus of object detection technique on basis of feature detection lies in discovering interest, regularly happened at the variations due to intensity, scale, brightness and a transformation that preserves co linearity and ratios of distances. Lower presented scale-invariant feature transform (SIFT) descriptor which is used in vision applications as a famous feature detection approach. It uses extrema as a part of scale space for determination with a different pyramid of Gaussian filters, and key points with low difference or inadequately restricted on an edge are evacuated. Next, a reliable introduction is allotted to each key point and its magnitude is evaluated on the basis of picture inclination histogram, in this manner accomplishing invariance to picture rotation. At each key point descriptor, the part of all picture inclination are inspected and weighted by a Gaussian, and after that denoted by orientation histograms.

For instance, the 16x16 specimen picture locale and 4x4 array of histograms with 8 alignment bins are regularly utilized, consequently giving a feature vector of dimension 128 for each key point. Various applications have been produced utilizing the SIFT descriptors, including object recovery [10] and object class revelation [3]. The Flow diagram of SIFT algorithm is shown in figure 5.

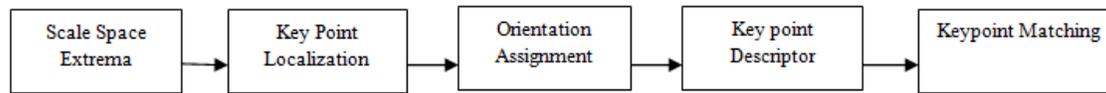

Figure 5: Flow of SIFT Algorithm

Talking about digitized world, head pose estimation is the method of detecting the direction/look and awareness of a human head through computerized method. It needs some steps to change a pixel-based representation of a head concerned to direction. The paper [11] illustrates the head pose estimation in order to identify the consciousness of driver/user by identifying his/her head orientation. This method takes advantage of symmetricity of the face to detect the head position. The proposed symmetrical strategy is independent of intrusion and can collaborate only with a specific user data. The method process in following way: At first the face is found out of the background image and then the symmetry of the face is detected. Later the picture is rolled/ rotated according to the need of inclination. Then the face features like eyes, nose. lips etc. is recognized through yaw featured approach [11].

Since most of the existing strategies work on 2D pattern object recognition, there is an expanding scope for techniques that unequivocally represent view variety and that consolidate recognition with pose estimation. In [7] a strategy is given that incorporates object detection and pose estimation utilizing 3D models and exhibits this methodology on the issue of car detection. The comparative study of various techniques for initial pose estimation techniques for monocular navigation has been done in [12].

Humans can observe and recognize the objects in any posture very easily. But mentioning about computing machines, specific problems arises due to change in view or pose. The issues of categorizing objects from multiple views and recovery through pose is combined, and straightforwardly affected by the way shape is described. Author exhibits a work structure that depends on homomorphic mapping between a typical and object. These manifolds with a specific end goal to mutually tackle the 3 problems of object recognition: category, instance and posture detection. Many trials on expansive information sets approve the quality of our methodology. Author has additionally demonstrated the ability of the methodology in evaluating full 3D posture. Author has also demonstrated the potential for real-time application to Artificial Intelligence and mechanical visual thinking by building a working closer to framework [13].

The work introduced in paper[14]is about mechanical robotization of the production line that is described by an extensive variety of items in concurrent assembling. The generation procedure restricts the utilization of traditional recognition frameworks joined to the object close by. Moreover, the mechanical structure of the conveyor presents geometric inaccuracies in the positioning. With the right categorization, the robot will have the capacity to self-governing choose the right program to execute and to perform coordinate framework rectifications. Perfect matching algorithm was given as the first solution for this problem. In this algorithm the 3D image of an object is compared with the known or predefined models. Perfect match algorithm gives high precision and makes the technique suitable for industrial applications. The whole identification and posture estimation strategy is performed in a scope of 3s with standard off the rack equipment. It is normal that this work adds to the incorporation of modern robots in exceptionally alterable and specialized production lines.

Another experimental recognition survey of object was done taking apples/fruits as object. The motive of this experiment was to automatically detect the actual defective fruit on the basis of its appearance because the concavity of the stem part also gives the same impression as defect. This recognition is done by near-infrared linear-array structured light. The 3D layout of the fruit is constructed and the upper half surface is reconstructed through multi-spectral camera. An inspected apple is taken as ideal reference for the further inspection of defect or stem. A complete spherical model is also taken as standard for observation and comparison with the apple and then a successful detection of concave stem is done. The various 3D impressions/images of apples are captured from different poses. After the completion of whole scanning process of upper part of the apple, the height of every pixel is measured and given as gray level. The proposed framework is very low cost and complex designing and its strategies can effectively recognize the stems from actual imperfections with high precision [15].

The use of RGB-D sensors for recognition of objects and posture orientation is studied in this paper [16]. Techniques are proposed on the basis of DAR methods i.e. depth-assisted rectification, which modifies characteristics separated from the colour picture utilizing depth information as a part of request to acquire a representation invariant to rotation, scale and point of view imperfections. The two techniques are DAR of patches and DAR of contours. It is seen that RGB-D sensors improves recognition of objects and posture orientation. The depth feature is used to represent scale-invariance and when DARP and DARC used together gives the best outcome [16].

The pose and alignment of an object an also be determined through RFID technology with RFID tags. Tags can be active or passive, but in [12] passive tags are mounted on each object. The RFID antenna then detects the tag on the object. This scheme works on the basis of power-map-matching strategy. Three tags are deployed on a single object. But here the power maps are created on basis of the basic mathematical model. This work is a low budget approach for estimating pose and alignment of an object. Experimental results show that the error (orientation/localization) is inversely proportional to tag separation. Thus it can be lessened with increase in tag separation [17].

Constant colour is a major feature of object detection approach, as colour changes moderately with respect to the brightness, posture of an object, and camera's view. These constant colour parameters are utilized to acquire shape descriptors. Constant shape is also an important feature of object recognition scheme, with respect to the shape, posture of an object, and camera's view. At that point, the colour and shape features are used together in colour-shape multidimensional which is later used as an index. For combining these both, various techniques are proposed till now. According to the indexing approach, it uses colour-shape connection to give a high-discriminative data which is strong against different image conditions. From this research it is seen that the detection of three-dimensional rigid/solid objects is more accurate and they are prune to variation in light, pose and view [18].

Another three-dimensional indexing scheme is also proposed in [19]. This method merges the feature-based techniques with alignment-based techniques. Here the research is concentrated on indexing models. LSH (Locality Sensitive Hashing) approach is used as a first step in order to determine the identical descriptor from the database. The important motive of LSH is to hash features into bins based on the collision probability. In second step, sampling of doublet matches is done. The pose of the object is detected using the matches similarity through joint-3D signature estimation. The goal of using these above mentioned approaches is to address the problem of vehicle detection. As the vehicle models are very similar, they are difficult to distinguish. Thus the proposed technique can meet this problem and has a capability rectify it [19].

In this article [20], a mutual context model is proposed which utilizes the object and human pose jointly by HOI (human-object-interaction) activity. This can be explained well as the method of detecting the object or human-pose separately which will later help the recognition of each other. This is useful in recognizing the player's pose or an artist's pose since because many times the image/video is captured from a distance. The human pose is many times occluded, which creates confusion in recognizing object and body parts. Like talking for a batsman's (player's) case, if we can recognize the bat in batsman's hand then the pose of the batsman is detected.

If the texture of an object is not perfectly defined, then it is very difficult to recognize the object and its posture. To address this problem, a new method of computing descriptors for object viewpoints is proposed. Here, the Nearest Neighbour searching method is implied for the situation where the objects and their poses are large in number. In order to implement this in reality, a neural network is convolutionally trained to employ similarity and difference among the descriptors. Wolhart et.al. states that their approach is capable to overcome occlusion and separate the different objects and poses of the objects from the given images. And depending upon these pose sets, the Euclidean distance (ED) varies accordingly i.e. if the poses are of different objects, the ED is high and if the poses are of same object, the ED is low. These features permit this approach to detect the actual object and actual pose [21].

To compute the perfect match for high dimensional vectors is known as Nearest Neighbour Matching (NNM). Due to the fast operation of NNM, the speed of various applications can be improved. The NNM approach and its aspects rely upon information set features and also define procurement for searching the suitable approach to find a specific dataset. The two algorithms are chosen for high-dimension-feature-matching: i) Randomized-k-d forest and ii) Priority search k-means tree [22].

Table 3: Comparison of Different Object Detection Methodologies their Accuracy and Error Rate

| Proposed Methodology | Error Rate | Accuracy |
|---|---|---|
| Generalized Gaussian kernel correlation (GKC)[23] | 3.56cm on SMMC-10 Dataset | 97.6% in Rigid Objects |
| 3D face Morphing with Depth Parameters[24] | 7.93 degree and 4.65 degree | 98.2% for continuously Varying Pose |
| Clustering of Pictorial Structure Model[25] | 1.23% of Error | 20% increase from previous method |
| Modified kernel density approximation [26] | Nil | 1.5 to 4.0% improvement over previous methods in finding correct localized parts |
| Local and contextual depth features[27] | Error rate with occlusion is reduced by 16% | 99.8% |

The Accuracy of different methodologies proposed in various papers is compared with the help of the line chart which is shown below. Four methodologies which is much famous is been compared with a lowest of 88% to the highest of 99% accuracy. This line chart helps in ease understanding of accuracy of various methods.

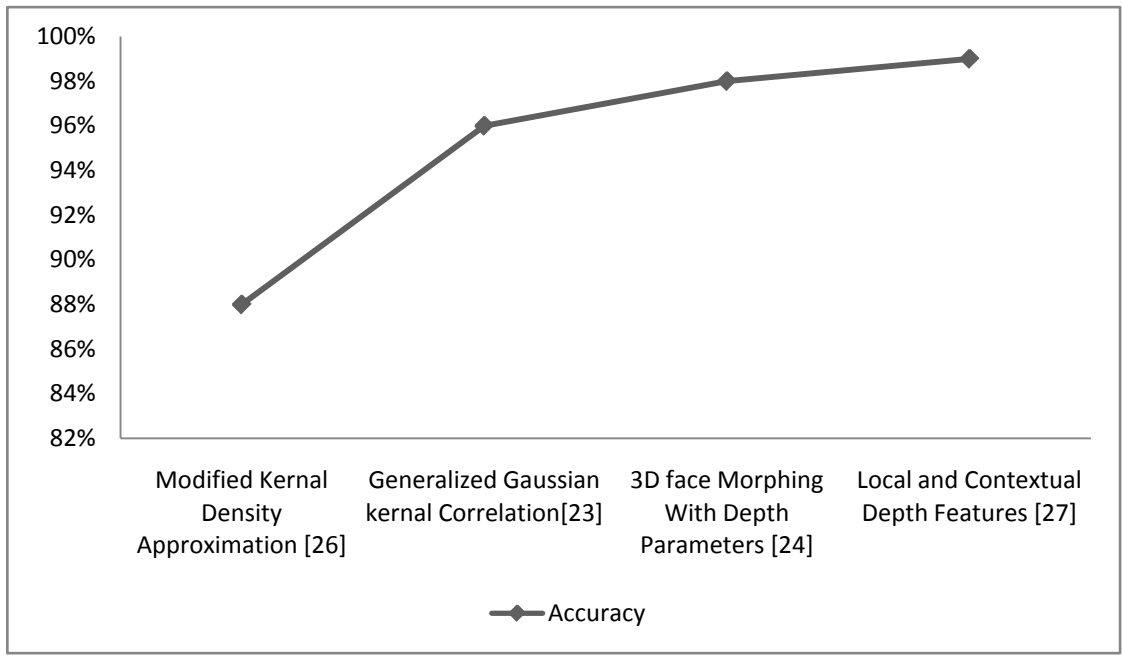

Figure 6: Comparison of Accuracy

## VI. Issues in Object Recognition

- The brightness of light may vary over the span of the day. No matter the light condition is the system must detect/identify the real object from the given image.
- In the given picture, the position of the object might be changed. The system must tackle such pictures consistently, if template matching is being utilized.
- The system must be efficient to handle the rotated image.
- The system must recognize the original picture if mirrored form of the image is given.
- The object recognition system must be able to detect the object which is not completely visible (i.e. hided or overlapped by other objects). This situation is known as occlusion.

## VII. Discussion

In [7] and [11], the head pose estimation issue is addressed. The prediction of yaw angles, pitch angles and roll angles is done on the basis of symmetry of the face. When the results were analyzed, it has shown a better performance of this approach as compared to state-of-art method. Various datasets were experimented and observed in [16] and shown that the proposed detector gives a 97.5% accurate pitch results and 98.2% accurate yaw results.

Experiment performing detection of different cars (taken as object) from distinct views/poses was discussed in paper [13]. Thus the viewpoint is estimated through voting. As the voting technique is described, it produces a set of recognitions in first stage to verify the object. In the second phase of this approach, the further processing of these applicants is handled. The bouncing box location is clarified. Every identification is scored that depicts the trust in it. Lastly, view estimation is practically rectified.

The survey work introduced in this paper [14] concentrates on the advancement of a 3D object limitation and detection framework to be utilized as a part of mechanical technology. The perfect match algorithm is suggested which proved itself worth by giving 99.5% accurate result when the classification of 8 object was performed by trying 200 its samples.

Another experimental recognition survey of object was done taking apples/fruits as object. It was a challenging task to compute the actual the difference among defects and stem of apples. Only single camera is used in combination with near-infrared lighting in this research [15]. When this research was experimented over 100 samples it has given 97.5% accurate results.

In [16], it is seen that RGB-D sensors improves recognition of objects and posture orientation. The depth-assisted method takes advantage of depth information and using the 3D data rectifies the patches so as to remove viewpoint impact.

## VIII. Conclusion

Most computer and robot systems have the ability to recognize the objects correctly. Though, there is wide improvement in this field, yet some issues/difficulties such as occlusion, rotation and change in brightness are being an obstacle in its progress. Various existing object recognition techniques and the problems faced are reviewed. Current techniques like 3D object recognition, face recognition, DAR method, head-pose estimation, driver head estimation etc. are already a part of many applications. But inspite of this great advancement object recognition is still not used in many areas. So still there is a scope for new object recognition classes that will prove worthy and will be considered for other application areas as well.